\documentclass[letterpaper, 10 pt, conference]{ieeeconf}  

\IEEEoverridecommandlockouts                              

\usepackage{graphics} 
\usepackage{epsfig} 
\usepackage{mathptmx} 
\usepackage{times} 
\usepackage{amsmath} 
\usepackage{amssymb}  
\usepackage{cuted}
\usepackage{float}
\usepackage{subfigure}
\usepackage[ruled,vlined]{algorithm2e}
\usepackage{ragged2e}

\title{\LARGE \bf
Automatic Construction of Lane-level HD Maps for Urban Scenes
}

\author{Yiyang Zhou$^{\ast}$$^{1}$ and Yuichi Takeda$^{\ast}$$^{2}$ and Masayoshi Tomizuka$^{1}$ and Wei Zhan$^{1}$
\thanks{$^{\ast}$Both authors contributed equally to this work. Yuichi Takeda completed this work during his visit at UC Berkeley.}
\thanks{$^{1}$Mechanical Systems Control Lab, University of California, Berkeley, CA, USA, 94705}
\thanks{$^{2}$Nissan Motor Co. Ltd., 1-1 Morinosatoaoyama, Atsugi,
Kanagawa, 243-0123, Japan}
\thanks{Correspondence: \texttt{yiyang.zhou@berkeley.edu}}
}

\begin{document}

\renewcommand\thefigure{\arabic{figure}}
\setcounter{figure}{1}

\maketitle
\thispagestyle{empty}
\pagestyle{empty}

\begin{strip}
\vspace{-25mm}
\centering\noindent
\begin{center}
\includegraphics[width=0.7\textwidth]{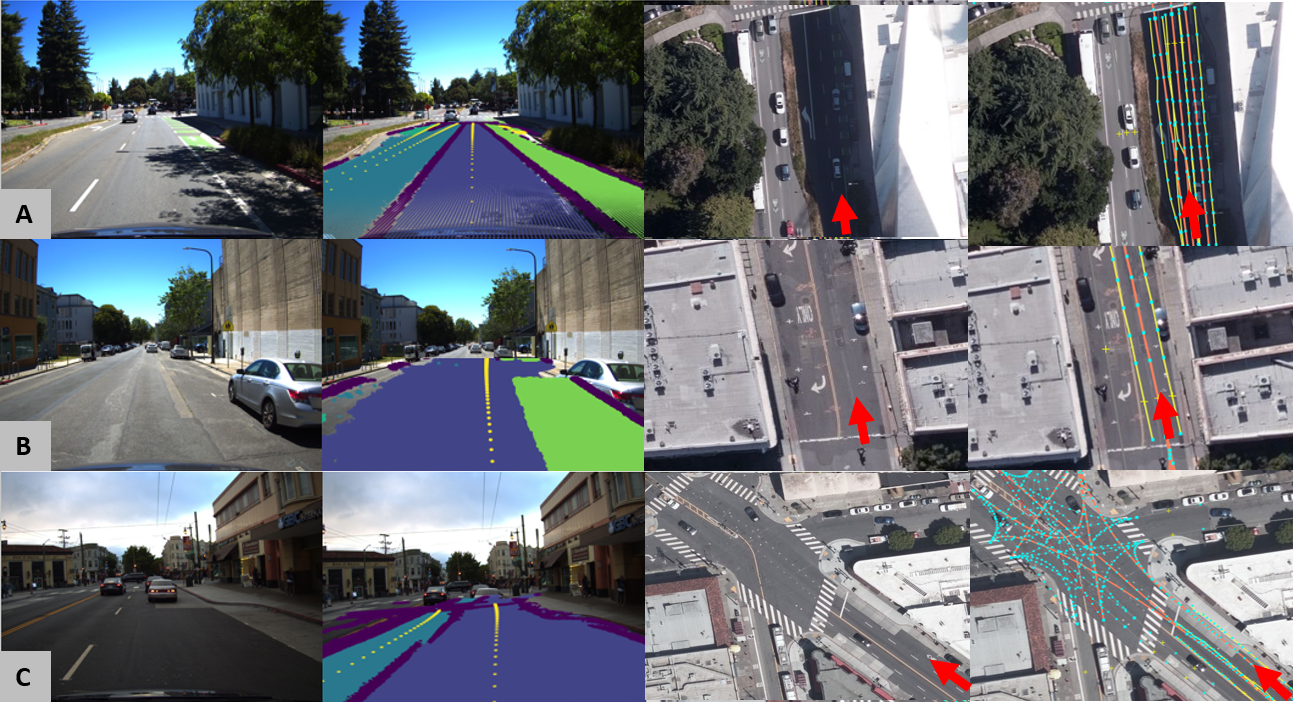}
\end{center}
\begin{justify}

\title{Fig.1 }{Fig.1 Roads in an urban environment are complicated: there are scenes with lane splitting (A), broken or no markings (B), and frequent irregularities like bus stops (C). Urban intersections are even worse: shown in C is a 6-way intersection in San Francisco. The proposed algorithm can explore the lane structure and intersection topology in such complex environments. The left two columns are camera inputs and generated HD maps overlapping the camera images: the aquamarine and blue areas represent left/center lanes; green represents other road elements like bike path or parking zones; and the purple and yellow dots are lane boundaries and reference trajectories for the ego vehicle. The right two columns are the satellite maps overlapped by the HD maps. Red arrows denote the corresponding ego vehicle pose.}

\end{justify}
\end{strip}

\begin{abstract}
High definition (HD) maps have demonstrated their essential roles in enabling full autonomy, especially in complex urban scenarios. As a crucial layer of the HD map, lane-level maps are particularly useful: they contain geometrical and topological information for both lanes and intersections. However, large scale construction of HD maps is limited by tedious human labeling and high maintenance costs, especially for urban scenarios with complicated road structures and irregular markings. This paper proposes an approach based on semantic-particle filter to tackle the automatic lane-level mapping problem in urban scenes. The map skeleton is firstly structured as a directed cyclic graph from online mapping database OpenStreetMap. Our proposed method then performs semantic segmentation on 2D front-view images from ego vehicles and explores the lane semantics on a birds-eye-view domain with true topographical projection. Exploiting OpenStreetMap, we further infer lane topology and reference trajectory at intersections with the aforementioned lane semantics. The proposed algorithm has been tested in densely urbanized areas, and the results demonstrate accurate and robust reconstruction of the lane-level HD map.  

\end{abstract}

\section{INTRODUCTION}

High definition (HD) maps have become a crucial component for full autonomy in a variety of complex scenarios. Encoded with accurate and comprehensive information of the static environment, HD maps can significantly facilitate perception, localization, prediction and planing \cite{SEIF2016159}. HD maps contain multiple layers of information abstractions, and the lane-level information plays the quintessential role in many applications. Embedded with lane geometries, road semantics, and connection topology, the lane layer can be utilized for defining potential region of interest (ROI) for some key modules in autonomous driving, including but not limited to object detection \cite{Yang2018HDNETEH}, regulating the lateral location of the ego vehicle \cite{ubermaploco,takedaloco}, and predicting behavior of other vehicles \cite{gao2020vectornet}. 

With numerous alluring applications, these lane-level HD maps, however, do not scale easily. Many of these HD maps are restricted to small scale environments due to the high costs in manual labeling and maintenance \cite{paz2020probabilistic}. Recently, a few commercial products have been launched to automatically map highways \cite{nvidiaDriveMap}, where the lanes are structured and the markings are clear. In urban scenes, however, the roads are much more complicated. As shown in the first and third column in Fig. 1, urban roads may have complicated forking, potholes, broken markings or even no markings at all. Furthermore, a city road also carries irregularities such as parking zones, bus curbs, and bike lanes. Also, the topological relationship of urban lanes are much more intricate at complicated intersections. As a result, the automatic lane-level mapping in urban areas remains as an open and challenging problem. 

It is worth to notice that the lane-level HD map construction problem is different from geometry tasks like lane detection or trajectory inference. HD map requires a semantic and topological understanding beyond the instance level, meaning that the map should contain logic connections with geometric information. As compared with simple lane boundary regression, HD map constructors further infer the merging/forking relationship between lanes; as compared with trajectory inference, HD maps further address the lane topological relationship. 

Previous works have regarded lane detection and intersection trajectory generation as separate problems, and most of these works only target on the geometric understanding of the scenes. In this work, we study lanes and intersections jointly and provide topological understanding beyond simple geometric representations. 

In this paper, we firstly define the HD map representation as a directed cyclic graph (DAG) for easy data storage and query. We then propose a method based on semantic-particle filter to automatically generate an urban lane-level HD map with a front view camera and an optional LIDAR sensor. The proposed method contains three major components: a semantic segmentation network for scene understanding, a sequential Monte Carlo lane tracing module over bird's-eye-view (BEV), and an intersection inference module with OpenStreetMap (OSM) \cite{OSM}. The whole pipeline only requires one single execution per road direction for a complete reconstruction of the lane-level details including the lane boundaries, reference trajectories, lane splitting information, and road topology. Lastly, we represent our generated lane map in a differentiable format for downstream modules. We test the proposed method in densely urbanized areas such as San Francisco and Downtown Berkeley from the UrbanLoco dataset \cite{wen2020urbanloco}, and some exemplar mapping results can be seen in Fig 1. The experiment covers areas with lane merging/splitting, missing/broken lane markings, complicated intersections, and irregular road shapes such as bus curbs and parking areas. The results demonstrate an accurate and robust construction of the lane-level HD maps. 

The major contributions of this paper are:
\begin{itemize}
    \item Propose to combine semantic scene understanding with Monte-Carlo sequential exploration for accurate and robust HD map construction in urban scenes.
    \item Infer geometrical representation and topological relationships for both lanes and intersections. 
    \item Exploit OSM as a coarse prior map, and construct a directed cyclic graph representation of the urban road structure. 
    \item Test the proposed algorithm with public dataset collected from densely urbanized area and validated the robustness and accuracy. 
\end{itemize}

\section{RELATED WORKS}
With the map-related research advancing in the past few years, there are a number of attempts for automatic lane-level map construction. We will start by introducing geometrical inference works for lanes and intersections. Later, we will introduce methods with topological understanding contributing to an HD map construction. 

\subsection{Lane detection: markings}
Since the lanes are defined by markings painted on the road surface, a natural way for lane-level map construction initiated from the lane marking detection perspective. Nieto et al. used a step-row filter \cite{steprow} for lane marking detection and applies Rao-Blackwellized Particle Filter for lane tracing \cite{NietoReal-timeFilter}. In \cite{Li2017DeepScene}, Li et al. used Convolutional Neural Network and Recurrent Neural Network for lane marking detection on highways. More recently, Garnett et al. \cite{Garnett20193D-LaneNet:Detection} and Guo et al. \cite{GuoGen-LaneNet:Detection} further used the 3D-LaneNet framework not only to classify the lane in an image, but also to predict its location in 3D. 

These approaches achieve high quality marking detection on highway or suburban roads where the shape of the roads is simple and can be approximated by a polynomial-like functions. However, it is hard to implement the aforementioned methods to urban scenarios with complicated road structures, broken/missing lane markings, and frequent road splits. Furthermore, the method mentioned in \cite{Garnett20193D-LaneNet:Detection} and \cite{GuoGen-LaneNet:Detection} only works for scenarios with mild changes in 3D slope, disregarding the abrupt changes in road topography. 

\subsection{Lane detection: drivable areas}
Some other methods focus on  drivable areas for lane detection. Meyer et al. designed a neural network for ego and neighboring lane detection \cite{Meyer2018DeepDriving}. Kunze et al. created a scene graph from semantic segmentation to generate a detailed scene representation of the drivable areas and all road signage \cite{Kunze2018ReadingScenes}. On the other hand, Roddick et al. used a pyramid projection network to extract the drivable area as well as other vehicles \cite{RoddickPredictSemanticMap}. Neither of these methods were extended to the HD mapping domain, allowing a third vehicle to make full use of the detection results. 

\subsection{Intersection lane inference}
Understanding the intersection structure is an indispensable technique for autonomous driving and HD Map generation. Thus a number of studies have been conducted to extract invisible lanes and connecting topology at intersections. 

Trajectories of other vehicles have been commonly used for intersection exploration. For example, in \cite{Geiger20143DPlatforms} \cite{Joshi2014JointStructure}, the authors used vehicle trajectories acquired from on-board sensor such as stereo camera or LIDAR. Later, Meyer et al. used simulated vehicle trajectories and employed a Markov chain Monte Carlo sampling to reconstruct the lane topology and geometry at both real and simulated intersections \cite{Meyer2019AnytimeParticipants, meyer2020fast}. In \cite{Roeth2016RoadMeasurements, Chen2010ProbabilisticTraces}, the authors used collected GPS data loaded on fleet vehicles. These methods have the potential to estimate lane-level structure of intersection, but they are data-hungry, and the performance heavily relies on the quality of vehicle trajectories, which themselves are non-trivial to acquire. 

More recently, directly predicting road connections at intersections from camera images became another popular direction for lane inference. The work from Nvidia \cite{lss} formulates the inference problem as classification and chose the best trajectories from a real trajectory pool with cross entropy loss. And Paz et al. \cite{paz2021tridentnet} predicts the trajectory from a Gated Recurrent Unit on a BEV semantic map. However, both methods only focus on the ego vehicle lane and predicts only one trajectory as a reference for the ego vehicle, not considering all visible lanes at an intersection. 

\subsection{HD map generation}
Large-scale map generation has long been considered as a Simultaneous Localization and Mapping (SLAM) problem \cite{Yang2018AMapping}. However, instead of semantics, common SLAM algorithms focus more on the odometry estimation. Recently, in \cite{paz2020probabilistic} and \cite{elhousni2020automatic}, the authors used semantics in the image domain and fuse semantic results with LIDARs to generate a semantic representation of the scene, not exactly an HD map. In terms of lane level map generation, a common approach is to accumulate extracted road environments along localization results \cite{Guo2016ASensors, Joshi2015GenerationLaneLevelMaps}. Both aforementioned approaches used LIDAR to extract road environments, and used OSM as a map prior. In \cite{Joshi2015GenerationLaneLevelMaps}, fork and merge of the lanes were able to be recognized by particle filter based lane marking tracking. Homayoundar et al. on the other hand, used a directed acyclic graph (DAG) and treated each detected lane marking as a node to decide future actions for connection, initiation, or termination \cite{Homayounfar2019DAGMapper}. However, these methods focused on road segments ignoring intersections, and they do not define the start and end point of the lanes. Thus, connections between lanes are not studied in these works. With the help of aerial images, authors of \cite{Mattyus_2016_CVPR} studied both road sections and intersections. However, the method is limited by the availability and resolution of aerial images. 

While the aforementioned efforts in lane detection and map generation has significant contributions to the community, they are limited to either simple road structures or single road segments. Our work targets specifically at the complicated urban driving scenes and considers both lane segments and intersection inference. Furthermore, we consider the topographical deformation of the road surface for more robust and accurate lane detection. 

\section{PRIMITIVE: MAP HIERARCHY}

\begin{figure}[t]
\centering
\subfigure[DAG Representation of the map $M$]
    {
\includegraphics[scale=0.27]{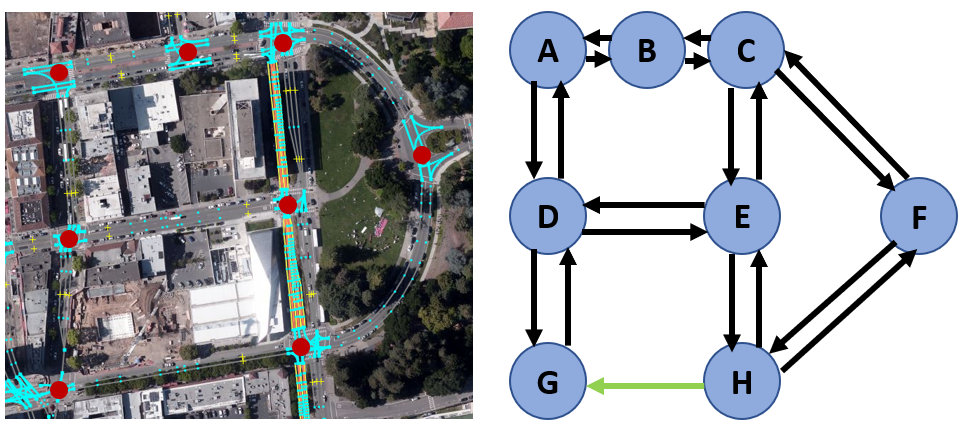}

    }
\subfigure[Intersection ROI]
    {
        \includegraphics[scale=0.20]{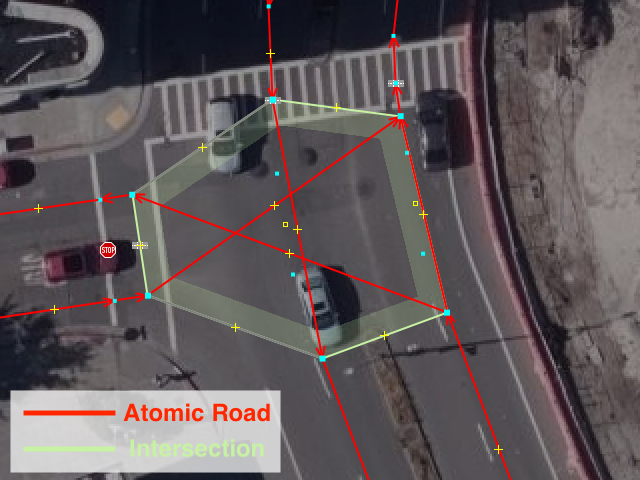}
    }
\caption{Map hierarchy as a directed cyclic graph. Each intersection (red dot) in the satellite map is represented by a node $I$ on the graph. The green edge $E$ represents an one way road.}
\label{fig:primitive}
\end{figure}

We depart from a topological point of view for this cartography task. A city road network is consisted of individual road segments and connecting intersections. As shown in Fig \ref{fig:primitive}-a, an urban lane-level HD Map $M$ could be further abstracted into a directional cyclic graph representation with each edge $E_{ij}$ representing a directional road from an intersection node $I_i$ to another intersection node $I_j$ . We deliberately choose to have an edge for each direction due to the ubiquity of one-way roads and physically separated two-way roads in urban scenes (as shown in Fig 1. A). Previous works focused on either the edge $E_{ij}$ or the node $I_i$ for geometrical information extraction, but we study the HD map $M$ containing both roads and intersections:

\begin{equation}
M:=\{E_{i\cdot},I_i\}
\end{equation}

To acquire such a high-level map skeleton, we extract the coarse road-level topological information from OSM \cite{OSM}. Disregarding the direction of travel for most roads, the OSM defines a road as a series of nodes connected together under a road instance. For special two-way roads with solid barrier as the median, OSM would have one road instance for each direction. Furthermore, the OSM defines an intersection as a connection node of two or more road instances. Although represented by a single node, the actual intersection is a geometrical region where multiple roads/lanes intersect. Utilizing such definition, we predict the intersection ROI (shown in Fig \ref{fig:primitive}-b) with a polygon formed by the closest road node to the intersection node, leaving the rest of the road nodes as part of the atomic road.

With the atomic road $E^{0}_{ij}$ and intersection patch $I^{0}_{i}$ defined as the skeleton $M^0$ of our map, we are able to match the moving data collection vehicle to an atomic road or an intersection in the road network. Different from the goal variables $M$, $E_{ij}$ and $I_{i}$, $M^0$, $E^{0}_{ij}$ and $I^{0}_{i}$ contains neither geometric nor topological information of the lanes. Now we proceed to the proposed methodology for automatic map generation. 

\section{METHOD}

\begin{figure*}[t]
\centering
\includegraphics[scale=0.5]{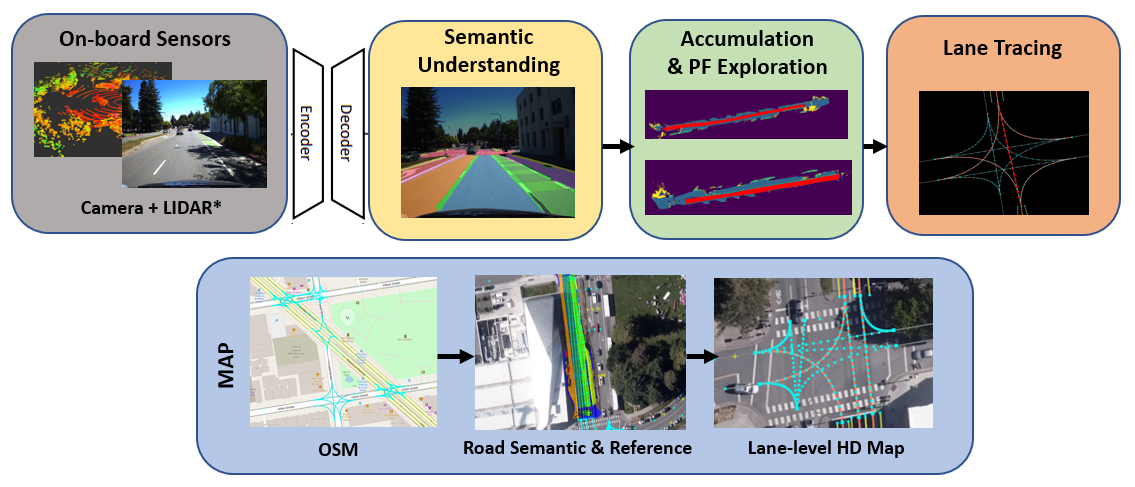}
\caption{The input of the proposed method is a series of camera images and LIDAR point clouds. We first process the camera image through a semantic segmentation module and then project the semantic instances onto the topographical mesh created by the LIDAR point cloud. We then accumulated the semantic projections from each frame to form a semantic map, in which a particle filter algorithm is carried out for lane regression. Lastly, we trace the lane connection at the intersection for topology.}
\label{fig:system_diagram}
\end{figure*}

\subsection{Problem formulation}
To automatically generate the lane-level map from the aforementioned map skeleton $M^0$, our proposed approach utilizes an on-board front-view camera $C_t$ and an optional (represented by $^*$) LIDAR $L^{*}_t$ at time $t$. We also require the synchronized vehicle poses $RT_t$ in the global coordinate, which come either from external sensors or SLAM algorithms. Representing our framework as a function $F(\cdot)$, the goal map shown in Equation 1 could be further described by:

\begin{equation}
M:=\{E_{i\cdot},I_i\} =F(C_t, L^{*}_t, RT_t, M^0)
\end{equation}

The goal for atomic roads $E_{ij}$ estimation is to infer lane $k$'s center line $\{L_k\}_K$ and its left-right boundaries $\{{B_{k,left}, B_{k,right}}\}_K$, where $K$ is the number of lanes in this atomic road. We deliberately go beyond a lane width as an asymmetrical representation of lane boundary due to irregularities in the drivable areas of a lane (examples are in Fig.1 B and C). Here, $L_k$ could be represented either as a continuous trajectory function and as a collection of way points, while discrete points ${B_{k,left}, B_{k,right}}$ would form an enclosed drivable area of the lane. As shown in Fig. \ref{fig:system_diagram}, we would start with semantic segmentation ($S(C_t)$) of the camera image, and then use particles to explore the lane over the BEV domain, which is accumulated from $S(C_t)$, $L^{*}_t$, and $RT_t$. To be more specific, we are looking for:

\begin{equation}
E_{ij}= \{L_k, B_{k,left}, B_{k,right}\}_K = G_1(S(C_t), L^{*}_t, RT_t)
\end{equation}

The goal for intersection $I_i$ estimation is to infer the topological relationship and geometrical reference trajectory between $E_{\cdot i}$ and $E_{i \cdot}$ as a Bezier curve $\mathbf{B}(E_{\cdot i}^k,E_{i \cdot}^l)$, where $k$ and $l$ denotes specific lanes in atomic roads. When two lanes are not topologically connected, the function output is set to be null. Here, we are utilizing the result in the previously defined $E_{ij}$ and the skeleton map $M^0$ for lane tracing at the intersection. Again, we study:

\begin{equation}
I_{i}= \{\mathbf{B}(E_{\cdot i}^k,E_{i \cdot}^l)\}_{K\times L} = G_2(E_{i \cdot}, M^0)
\end{equation}

As an outline for the following sections, function $S(.)$ is introduced in Section IV-B, function $G_1(.)$ is introduced in Section IV-C, and function $G_2(.)$ introduced in Section IV-D.

In corresponding maps shown in the bottom of Fig. \ref{fig:system_diagram}, we start with a coarse map skeleton $M^0$, go through the road semantics $E_{ij}$ and reference trajectory $I_{i}$ generation, and end with a lane-level HD map $M$. 

\subsection{Semantic understanding of the scene}
Drivable areas and lanes are defined by the road markings painted on road surfaces, and lane markings in the urban areas are more complicated: there are numerous lane splitting, frequent stop lines and broken/missing markings. 

\begin{figure}
    \centering
    \subfigure
    {
        \includegraphics[width=1.56in]{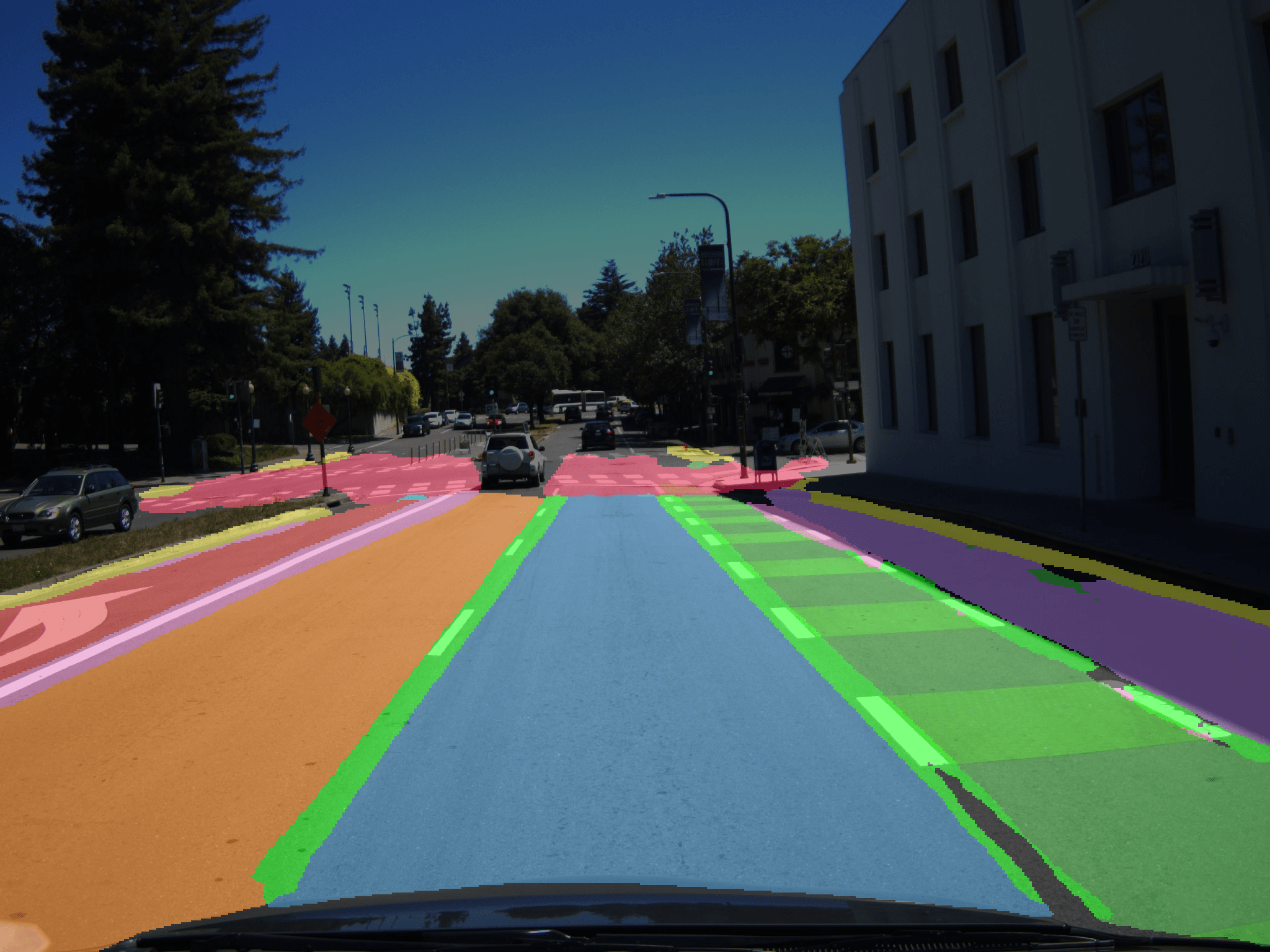}
    }
    \subfigure
    {
        \includegraphics[width=1.56in]{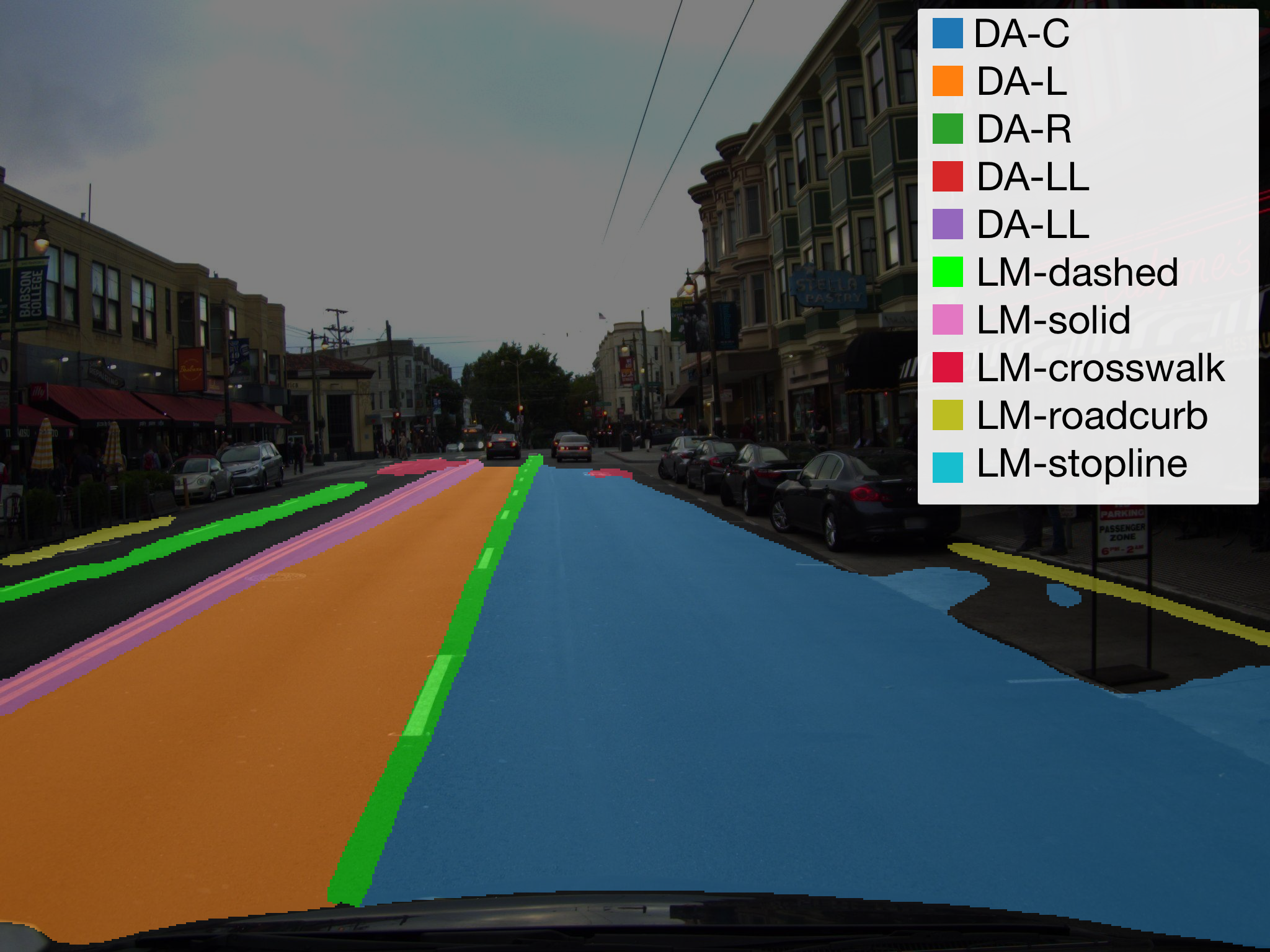}
    }
    \caption{Example results of semantic understanding module. This module extracts not only the lane marking but also the lane itself. DA- represents Drivable Area, which is Center, Left, Right, LeftLeft, RightRight respectively. LM- is a prefix of road marking class.}
    \label{fig:semantic_segmentation}
\end{figure}

Therefore, instead of only extracting lane markings from camera images, we consider this problem as semantic segmentation and infer both drivable areas and lane markings from camera images. Built upon a DeepLab-v3+ \cite{Chen2018Encoder-decoderSegmentation} structure, we are particularly interested in ego lanes, 2 neighboring lanes on each side of the ego lane, dashed lines, solid lines, crosswalks, road curbs, and stop lines. As shown in Fig. \ref{fig:semantic_segmentation}, the input of the network is an image $C_t$, and we predict the aforementioned 10 classes. These semantic instances forms the foundation of our understanding of the current scene. Considering both lanes and lane markings is especially efficient for urban scenes as the lane markings might be missing, and some drivable areas may be confused with parking zones. 

\subsection{BEV accumulation and atomic road structure estimation}
After the road semantics are extracted from the camera images, the segmented images ($S(C_t)$) are projected onto the ground plane in map coordinates. These projected BEV images are then accumulated together for a semantic representation of the atomic road. For the scope of this semantic mapping problem, we assume that the pose of the ego vehicle $RT_t$ is given from auxiliary sensors like inertia-GPS navigation systems or SLAM algorithms like \cite{Yang2018AMapping}. 

\begin{figure}
    \centering
    \subfigure
    {
        \includegraphics[width=1.0in]{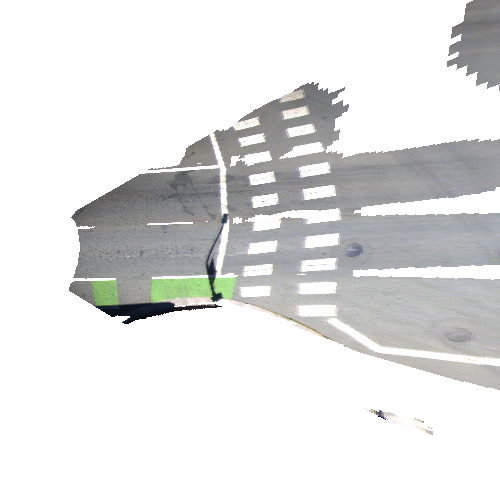}
    }
    \subfigure
    {
        \includegraphics[width=1.0in]{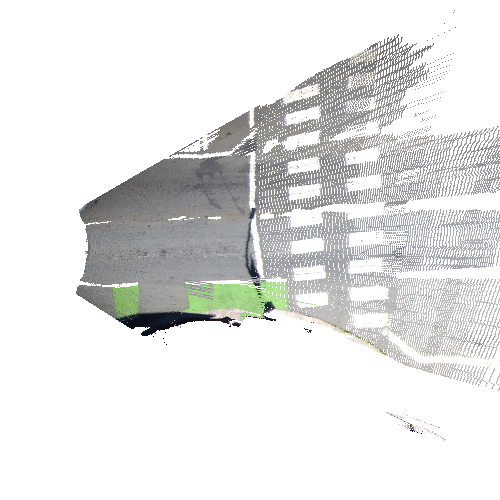}
    } 
    \caption{BEV projection comparison with LIDAR (left) and without LIDAR (right). Without LIDAR, the crosswalk is significantly distorted.}
    \label{fig:bev_crosswalk_comparison}
\end{figure}

For the BEV projection, most previous works assume a flat ground as the road model. However, such estimation is over-simplified in city scenes. For urban areas, undulating road surfaces, either purposefully designed for drainage or accidentally caused by lack of maintenance, are common. Fig. \ref{fig:bev_crosswalk_comparison}-left shows that by simply assuming a flat surface of the road, the BEV projection could be distorted. We propose to optionally use a sychronized LIDAR scan $L_{t}^*$ as the ground topography, and then project the image on to a topography mesh. To generate the ground mesh from sparse point clouds, we process the synchronized LIDAR point cloud through a Delaunay Triangulation with each point as a vertex on the mesh. After LIDAR correction, the BEV projection could be seen in Fig. \ref{fig:bev_crosswalk_comparison}-right, where the distorted cross-walk markings are rectified. To further demonstrate the improvement of using the true ground topography, we include an ablation study of the map quality with/without the LIDAR correction in Section V-C.

\begin{figure}
    \centering
    \subfigure[]
    {
        \includegraphics[height=1.4in]{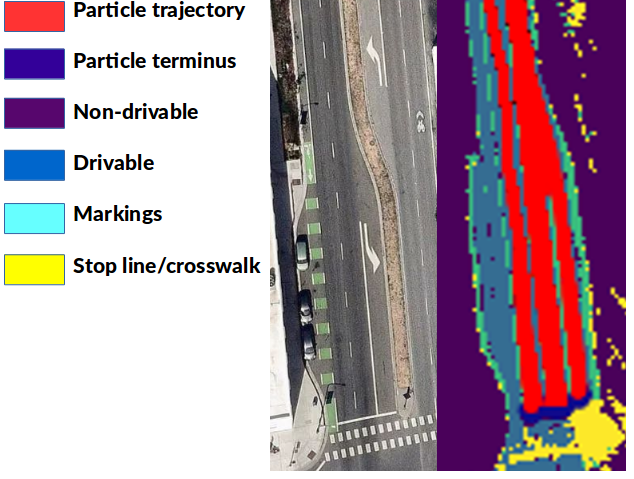}
    }
    \subfigure[]
    {
        \includegraphics[height=1.4in]{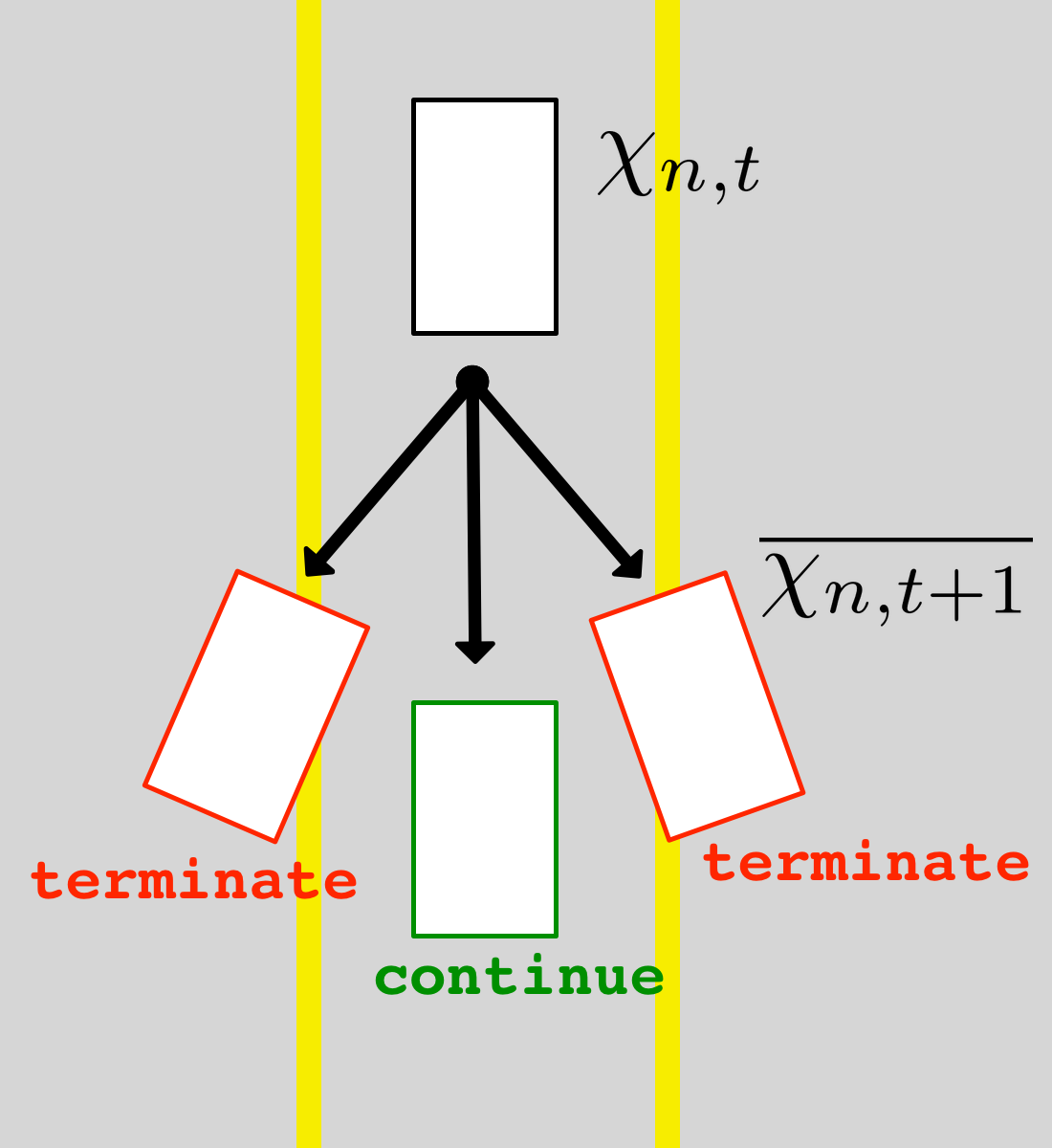}
    }
    
    \caption{(a): Particle history plotted next to the satellite map. The particles were exploring the forking at this atomic road. (b): Prediction and evaluation step of particle filter. Particle at time $t$ ($\chi_{n,t}$) evolve by Eq. \ref{eq:pf}. Each predicted particle $\chi_{n,t+1}$ will be evaluated by BEV semantic map. If the particle overlaps with lane boundary, it will be terminated, and the only particles doesn't touch will continue to next step.}
    \label{fig:bev_atomic_road_comparison}
\end{figure}

With a semantic atomic road map shown as the background of \ref{fig:bev_atomic_road_comparison}-a, we now study the possible vehicle trajectories in such scenarios. To tackle the complicated, mostly irregular, driving scenes, we propose to use a Monte Carlo exploration strategy: the particle filter. Each particle represents a moving vehicle of an average sedan size with three state parameters: the BEV location and the yaw angle: $\chi_{n,t}= [x,y,\phi]_{n,t}^{T}$. The details of this exploration strategy is shown in Algorithm 1. With the ego vehicle starting at one end of the atomic road with $RT_0$, we generate a strip of $N$ particles $\{\chi_{n,t=0}\}_N$ perpendicular to the driving direction in $RT_0$. Each particle $\chi_{n,t}$ then proceeds along the driving direction in the atomic lane map shown in Fig. \ref{fig:bev_atomic_road_comparison}-a. Here, we are simulating the actual driving of the car with speed $v_m$ and yaw-rate $\omega_m$ uniformly sampled from noisy linear and angular velocity distribution $V_m$, $\Omega_m$. The dynamic update function is

\begin{equation}
\label{eq:pf}
\overline{\chi_{n,t+1}}= \chi_{n,t} + \begin{bmatrix}
\cos{(\phi_{n,t}+\omega_m)}*v_m \\
\sin{(\phi_{n,t}+\omega_m})*v_m \\
\omega_m
\end{bmatrix} 
\Delta t
\end{equation}

Each predicted particle $\overline{\chi_{n,t+1}}$ will be re-weighted for it's overlapping ratio with the lane boundaries (Fig.  \ref{fig:bev_atomic_road_comparison}-b), and the weight is saved as $w_{n,t+1}$. Particle will be terminated at $I_j$. Through the sequential Monte Carlo process, we could simulate vehicle behaviors at lane splitting scenes. For unstructured roads too narrow to fit two lanes, particle filter also demonstrated a preferred driving trajectory of the vehicle (Fig.1 B). 

For particles travelled to the end of the atomic road $\{\chi_{n,T}\}$, we first clustered these particles with geometrical clustering algorithm DBSCAN \cite{DBSCAN} to determine the resulting number of lanes. Then we performed numerical regression to find the best representation of these lanes. After experimenting different regression models, we concluded that the best fit would be a optimized piece-wise linear regression smoothed with natural spline. By minimizing the sum-of-squared regression loss shown in Equation 6, we first determined the optimal break points $b_i$ for the piece-wise regression function $f(x_n,b_i)$, and then used the natural spline to smooth the connection for a differentiable curve representation. Such differentiable form of reference trajectory is particularly valuable for further path-planning and prediction tasks \cite{gao2020vectornet}. 

\begin{equation}
L_k= f(x_n, \underset{b_i}{argmin}( \sum_{t=0}^{T}\sum_{n=1}^{N} (y_n -f(x_n,{b_i}))^2))
\end{equation}

Sampling way points for each regressed lanes $L_k$, we estimate the lane width by probing to the latitudinal direction of the lane. As a result, we would get accurate lane width ${B_{k,left}, B_{k,right}}$ at each specific sampled location. Together we form the atomic road map $E_{ij}$ introduced in Equation 3 with reference trajectories $L_k$ their lane boundaries ${B_{k,left}, B_{k,right}}$. It is worth to notice that such representation could be easily transferred to a LaneLet2 \cite{lanelet2} format for further tasks. 

\begin{algorithm}
\SetAlgoLined
\KwResult{Returns the regressed lane on atomic roads on $E_{ij}$}
 initialize $N$ particles $\{\chi_{n,t=0}\}_N$ with $RT_{0}$ and size;\\
 \While{$\chi_{n,t}$ not at $I_j$}{
 $\{\overline{\chi_{n,t+1}}\}_N$, $\{{w_{n}}\}_N$ = $\emptyset$;\\
     \For{$n$ in 1 to $N$}{
        sample $\overline{\chi_{n,t+1}}$ from Equation 5 with $V_n$ and $\Omega_n$;\\
        $w_{n}$ = evaluate $\overline{\chi_{n,t+1}}$;\\
        $\{\overline{\chi_{n,t+1}}\}_N$ $\leftarrow$ $\overline{\chi_{n,t+1}}$;
        $\{{w_{n}}\}_N$ $\leftarrow$ $w_{n}$;
    }
    \For{$n$ in 1:$N$}{
        draw i from $\{w_{n}\}$;\\
        $\{\chi_{t+1}\}$ $\leftarrow$ $\chi_{i,t+1}$;\\
        save particle history;
}
}
cluster $\{\chi_{n,T}\}$ in $C$ with DBSCAN;\\
\For{$c_k$ in $C$}{
$L_k= f(x_k, b_k | x_k \in c_k)$\\
${B_{k,left}, B_{k,right}}$ $\leftarrow$ Longitude explore $L_k$}
\caption{Lane Exploration Particle Filter}
\end{algorithm}

\subsection{Intersection lane tracing}
\begin{figure}[tb]
\centering
\includegraphics[scale=0.12]{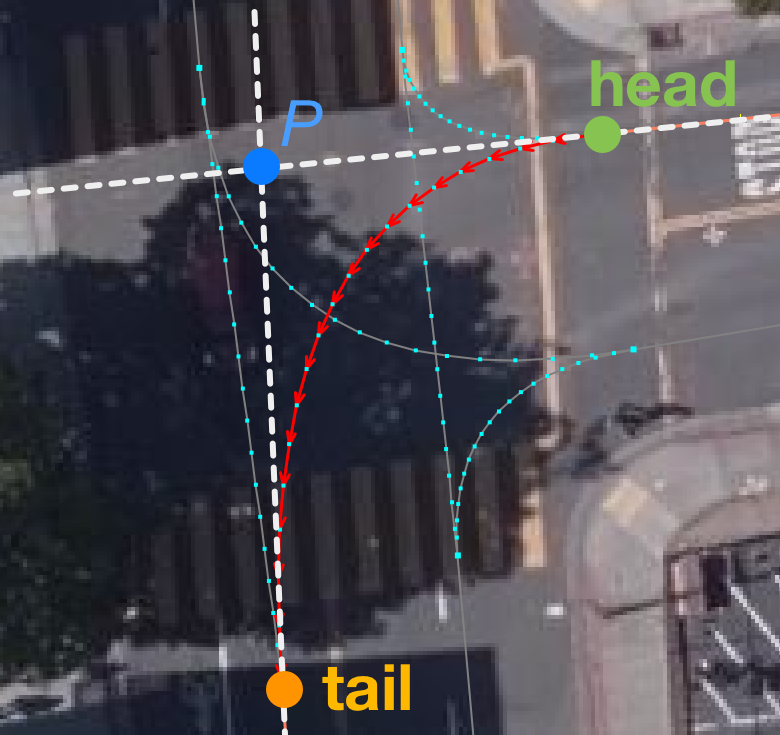}
\caption{Lane inference in intersection. We use second-order Bezier curve for lane geometry inference in intersection. The Bezier curve has 3 control points. Head and tail comes from last and first node of connected lanes, and the intermediate control point $P$ is a intersecting point of lines extending the direction from head and tail nodes. The red line represents inferenced lane center, and gray lines with blue dots are other lanes in this intersection.}
\label{fig:intersection_inference}
\end{figure}

As previously introduced in Equation 4, the road connection topology and reference trajectories inferences inside an intersection $I_i$ come from the OSM skeleton $M^0$ and the lanes generated on neighboring atomic roads $E_{\cdot i}$ and $E_{i \cdot}$. Since the OSM defines the road connecting topology, we transfer this relationship to the lane level with general traffic rules: i.e. left lanes in $E_{\cdot i}$ would connect to left lanes in $E_{i \cdot}$; and the algorithm is demonstrated in Algorithm \ref{algo:intersection}. Without explicit supervision, the topological inference would reach over 90\% precision in urban areas. Based on the topology, reference trajectories are then regressed with a optimized second-order Bezier curve shown in Equation 7. We optimize Equation 7 for smoothness over the variable $\alpha \in [0,1]$. The Bezier curve's head and tail correspond to the lane's location $L_{k_1,-1}$ and $L_{k_2,0}$ as shown in Fig. \ref{fig:intersection_inference}, and we used intersecting point of lines extending along the direction from the head and tail nodes as control point $\mathbf{P}$ for the curve.

\begin{equation}
\mathbf{B}(E^{k_1}_{ji},E^{k_2}_{il}, \alpha) =  (1 - \alpha)^{2}L_{k_1,-1} + 2\alpha(1 - \alpha)\mathbf{P} + \alpha^{2}L_{k_2,0}
\end{equation}
 
\begin{algorithm}
\label{algo:intersection}
\SetAlgoLined

\KwData{ $E_{ji}, E_{il}$: set of incoming and outgoing atomic roads to intersection $I^0_{i}$}
\KwResult{returns set of lanes $I_i$ for intersection $I^0_{i}$}

connections = $\phi$ \\
\For{$E_{in}$ in $E_{ji}$}{
\For{$E_{out}$ in $E^0_{il}$}{
$N_{in} = num\_lanes(E_{in})$ \\
$N_{out} = num\_lanes(E_{out})$ \\

\For{$k$ in 1 to $min(N_{in}, N_{out})$}{
$connections = connections \bigcup \{(E^k_{in}, E^k_{out})\}$
}
}
}

$I_i = \phi$ \\

\For{$\{(E^k_{in}, E^k_{out})\}$ in connections }{

$\mathbf{B}_k =$ Bezier curve from Equation 7 \\
$I_i = I_i \bigcup \mathbf{B}_k $
}

return $I_i$
\caption{Lane inference in intersection}
\end{algorithm}

\section{EXPERIMENTS}

\begin{table*}[ht!]
\begin{center}
\caption{Atomic Road Mapping Evaluation}
\vspace{2mm}
\begin{tabular}{|c|c|c|c|c|c|c|c|}
\hline
Approaches & Location & Urbanization & Sensor & \multicolumn{4}{c|}{Reported Results}  \\
\cline{5-8}
 & Name  & Rate& Input &{RMS Error (m) } & {mIOU} & Precision & Recall  \\
\hline
\hline
&Downtown  &  Dense&LIDAR+Cam & {$0.24$} & {$0.79$} & {$0.84$} & {$0.73$} \\

\textbf{Ours} & Berkeley &   &Cam & {$0.25$} & {$0.73$} & {$0.64$} & {$0.56$} \\

\cline{2-8}
& San Francisco&  Dense &LIDAR+Cam & {$0.33$} & {$0.76$} & {$0.63$} & {$0.63$} \\

\hline
\hline
Mattyus et al. \cite{Mattyus_2016_CVPR} & Karlsruhe & Low & Aerial+Cam & {$0.57$} & {$0.55$} & {$0.86$} & {$0.60$}\\
\hline
Meyer et al. \cite{Meyer2018DeepDriving}& Frankfurt &  High & Stereo & {$--$} & {$0.58$} & {$--$} & {$--$} \\
\hline
Paz et al. \cite{paz2020probabilistic}&San Diego, CA&  Medium & LIDAR+Cam & {$--$} & {$0.71$} & {$0.78$} & {$--$} \\
\hline
Joshi et al. \cite{Joshi2015GenerationLaneLevelMaps}&King, MI&  Low & LIDAR & {$0.06$} & {$--$} & {$--$} & {$--$} \\
\hline
Elhousni et al. \cite{elhousni2020automatic}& Not Known &  Medium & LIDAR+Cam & {$0.30$} & {$--$} & {$--$} & {$--$} \\

\hline
\end{tabular}
\end{center}
\vspace{-2mm}
\end{table*}

\begin{table*}[ht!]
\begin{center}
\caption{Intersection Inference Evaluation}
\vspace{2mm}
\begin{tabular}{|c|c|c|c|c|c|c|}
\hline
Approaches & Location & Urbanization &  Sensor & \multicolumn{3}{c|}{Evaluation Metrics}  \\
\cline{5-7}
 & Name &Rate  & Input & Precision & Recall & RMS Error (m)  \\
\hline
\hline
& Downtown  & Dense&  LIDAR+Cam & {$0.91$} &{$0.80$} & {$0.24$}\\
\textbf{Ours}& Berkeley & Dense&  Cam & {$0.90$}& {$0.81$} & {$0.30$}\\
\cline{2-7}
& San Francisco & Dense&  LIDAR+Cam  & {$0.93$} & {$0.90$}& {$0.46$}\\
\hline
\hline
Geiger et al \cite{Geiger20143DPlatforms}& Karlsruhe & Medium&  Stereo & {$--$} & {$0.92$}& {$3.00$}\\
\hline
Joshi et al \cite{Joshi2014JointStructure}& King, MI & Low&  LIDAR  & {$--$} & {$0.96$}& {$0.5$}\\
\hline
Meyer et al \cite{Meyer2019AnytimeParticipants}& Karlsruhe & Low&  Simulation  & {$--$} & {$0.85$}& {$0.27$}\\
\hline
Roeth et al \cite{Roeth2016RoadMeasurements}& Not Known & Mdeium &Fleet DGPS  & {$--$} & {$--$}& {$5.2$}\\
\hline
\end{tabular}
\end{center}
\vspace{-1mm}
\end{table*}

\subsection{Experimental Setup}
To validate the proposed method, we have created lane-level HD maps for 3.5 KM routes in 47 blocks of the San Francisco Bay Area. The data is from a public mapping dataset \cite{wen2020urbanloco} with front-view camera, LIDAR, and ground-truth ego-vehicle pose. To evaluate the atomic road and intersection reconstruction IOU, we manually label the drivable areas on Bing \cite{bingmap}, and we performed a global rectification \cite{TUM_ATE} for alignment as the maps and the ground truth were generated in different coordinate systems. To quantify the trajectory deviation, we use the trajectory from another driver passing through the same scenes on a different day. 

For the semantic understanding module, We adopt DeepLab-v3+ \cite{Chen2018Encoder-decoderSegmentation}.
We pre-train the module with the Cityscapes dataset \cite{cordts2016cityscapes} and fine-tuned with the BDD100K dataset \cite{yu2020bdd100k}. Since BDD100K dataset only defines ego-lane and alter lanes for drivable areas, we extend this label to ego-lane, left lane, two left lane, right lane, and two right lanes. 
We use SDG as optimizer with learning rate 0.01, momentum 0.9, and batch size 4. Fine tuning is done for 250,000 steps.
Images are resized to $512 \times 512$ on semantic segmentation process and restored to original size on BEV projection.

For BEV accumulation, we use the ground-truth localization provided in \cite{wen2020urbanloco}. In the particle filter exploration, we deployed 500 particles, each with a length of 3m and a width of 1.5m. The velocity (m/s) is drawn uniformly from $U(0.9,1.1)$, and the yaw angle (radient) is uniformly drawn from $U(-0.2,0.2)$. The maximum cluster distance for DBSCAN is set to be 1m, which is around half of a typical lane with in urban scenes.  
\subsection{Lane-level map in Berkeley and San Francisco}

A qualitative atomic road mapping result is shown in Fig. 1: the proposed method could successfully map lane-splitting (A), un-structured roads (B), and complicated intersections (D). For quantitative studies on lane-level HD maps, we do not find a consensus in academia for map quality evaluation. However, we use all popular metrics in previous works to demonstrate the effectiveness of our algorithm: the root-mean-square (RMS) error between the reference trajectory and ground truth, the mean intersection of union (mIOU) index of the proposed lane boundary, and the precision-recall values. All the evaluations are performed on per-lane basis. To study the detection rate of the proposed approach, we define a successful detection as one having over 0.7 mean IOU or less than 0.2m of RMS. A quantitative study of the proposed methods could be seen in Table 1. 

Gauged at 0.7 mIOU or 0.2m RMS, our proposed methods has a mean RMS of 24cm for lane center trajectory estimation, and an average IOU of 0.79 for lane boundary estimation in Downtown Berkeley. With the same threshold, we are able to achieve 0.84 in precision and 0.73 in recall. In North Beach, San Francisco, our approach has a RMS of 0.33m and mIOU of 0.76 when gauged at 0.7 IOU with 0.63 precision and 0.63 recall. 

To the best of our knowledge, most previous endeavors in the similar filed use private data for evaluation, and we could not test our proposed approach on their dataset. More critically, most mapping-related algorithms and parameters are close-sourced, making it impossible to re-evaluate on our dataset. Thus, Table 1 lists the performance of other proposed methods \cite{Joshi2015GenerationLaneLevelMaps, elhousni2020automatic,paz2020probabilistic,Mattyus_2016_CVPR, Meyer2018DeepDriving} in similar matrices. We also list the urbanization rate at each location for a qualitative review of the difficulties for mapping at these areas. 

For the invisible topology and trajectory inference in intersections, we evaluate our performance by the topological relationship precision-recall index as well as the inference trajectory RMS error. A quantitative results for the intersection could be seen in Table 2. Compared with a series of other methods in less urbanized areas, our algorithm is still able to discover the potential topological relationship at intersections with a lower RMS error.

\subsection{Ablation study: true road topography}
To study the effect of the road topography on our mapping system, we compare the map generated with true road topography and the map generated with a plane assumption of the ground. A quantitative comparison is shown in Table 1 and 2. It is clear that, with the LIDAR corrected topography, we have a significant improvement in atomic road detection precision and recall score. Also the estimated drivable area are closer to our labelled ground truth.

\section{CONCLUSION and FUTURE WORKS}
We have demonstrated that an HD map could be constructed automatically in complicated urban scenes with a well-designed framework. Semantic understanding and Monte Carlo exploration strategy form the core of our proposed method. By employing particle filters on the accumulated semantic BEV maps, we are able to discover the atomic road structures. Furthermore, by exploiting OSM, we inferred both topological and geometrical connections at intersections. Tested in densely urbanized areas, our proposed approach enables large-scale HD map constructions, further facilitating downstream modules for full autonomy. In the current pipeline, semantic segmentation on 2D images limited our capacity at heavily occluded scenes. For next step, we plan to employ amodal prediction pipelines to directly predict semantic maps in the BEV domain. 


\bibliographystyle{ieeetr}
\bibliography{automap_references.bib}

\end{document}